\documentclass[runningheads]{llncs}
\usepackage[T1]{fontenc}
\usepackage[figuresright]{rotating}
\usepackage{mathtools}
\usepackage{xcolor}  
\usepackage{graphicx}
\usepackage{bm}
\usepackage{makecell}
\usepackage{bbm}
\usepackage{xspace}
\usepackage{float}
\usepackage{color}
\usepackage{colortbl}
\usepackage{dsfont}
\usepackage{multirow}
\usepackage{multicol}
\usepackage{pgfplots}
\usepackage{algorithm}
\usepackage{algpseudocode}
\usepackage{adjustbox}
\usepackage{listings}
\usepackage{bbding}
\usepackage{pifont}
\usepackage{wasysym}
\usepackage{amssymb}
\usepackage{tabularx}
\usepackage[misc]{ifsym}
\usepackage{comment}
\usepackage{multirow}
\usepackage{colortbl}
\usepackage{pifont}
\usepackage{svg}

\begin{document}
\title{ICDAR 2025 Competition on End-to-End Document Image Machine Translation Towards Complex Layouts}

\author{ Yaping Zhang\inst{1,2}\orcidID{0000-0001-6892-905X} , Yupu Liang\inst{1,2}\orcidID{0009-0004-6121-9054}, Zhiyang Zhang\inst{1,2}\orcidID{0009-0004-4057-7098}, Zhiyuan Chen\inst{1,2}\orcidID{0009-0001-6921-6251}, Lu Xiang\inst{1,2}\orcidID{0000-0002-7702-7274}, Yang Zhao\inst{1,2}\orcidID{0000-0003-1028-3406},  Yu Zhou\inst{2,3}\orcidID{0000-0002-4911-4717}\textsuperscript{(\Letter)}, Chengqing Zong\inst{1,2}\orcidID{0000-0002-9864-3818}\textsuperscript{(\Letter)}}
\institute{Institute of Automation, Chinese Academy of Sciences;  
\and University of Chinese Academy of Sciences, Beijing, China; \\ 
\and Fanyu AI Laboratory, Zhongke Fanyu Technology Company Ltd., Beijing, China.
\email{\{yaping.zhang,lu.xiang, yang.zhao, yzhou,cqzong\}@nlpr.ia.ac.cn}\\
 \email{\{liangyupu2021,zhangzhiyang2020,chenzhiyuan2023\}@ia.ac.cn}}

%



\authorrunning{Y. Zhang et al.}

\maketitle             

\begin{abstract}
Document Image Machine Translation (DIMT) seeks to translate text embedded in document images from one language to another by jointly modeling both textual content and page layout—bridging optical character recognition (OCR) and natural language processing (NLP). 
The DIMT 2025 Challenge advances research on end-to-end document image translation, a rapidly evolving area within multimodal document understanding. 
The competition features two tracks—OCR-free and OCR-based—each with two subtasks for small ($\leq$ 1B parameters) and large ($>$ 1B parameters) models.
Participants submit a single unified DIMT system, with the option to incorporate provided OCR transcripts. Running from December 10, 2024 to April 20, 2025, 
the competition attracted 69 teams and 27 valid submissions in total. Track 1 had 34 teams and 13 valid submissions, while Track 2 had 35 teams and 14 valid submissions.
In this report, we present the challenge motivation, dataset construction, task definitions, evaluation protocol, and a summary of results. Our analysis shows that large-model approaches establish a promising new paradigm for translating complex-layout document images and highlight substantial opportunities for future research. 
\end{abstract}

\section{Introduction}
\quad Recent advancements in large-scale language models (LLMs) have led to significant breakthroughs in  optical character recognition (OCR)~\cite{dong2024internlm,yao2024minicpm} and machine translation of plain text~\cite{yang2025implicit}. 
However, the translation of real-world document images, characterized by intricate layout structures—including mixed column formats, tables, and footnotes—continues to pose substantial challenges for current LLM architectures~\cite{qian-etal-2024-anytrans,lan2023exploring,zhiyangTpamiUnidit25,liang-etal-2024-document}. This inherent difficulty significantly curtails the effective application of LLMs for comprehensive knowledge extraction and limits their broader utility in diverse document-centric applications.

The task of Document Image Machine Translation (DIMT) addresses this by aiming to translate textual content embedded within document images from a source language to a target language. DIMT is fundamentally a multi-modal problem, necessitating the sophisticated integration of visual layout understanding with natural language processing (NLP) capabilities, thereby bridging the traditional divide between OCR and downstream textual analysis. While recent years have witnessed notable progress in DIMT technologies \cite{zhang2023layoutdit,liang2024survey,liu2024ocrbench,donut,zhang2024llava}, existing solutions often fall short of the robustness and accuracy required for practical, real-world deployment. The primary impediments can be attributed to:

\begin{itemize}
    \item Multi-modality and cross-linguality: Due to the inherent multi-modality nature, real-world document images often involve a intricate combination of complex layout, dense text, and visually-rich elements, resulting in difficulties in their comprehensive understanding and the cross-lingual translation.
    \item Image and text noise: Many factors such as image defect or OCR error may cause noise to the image or text as model input, leading to additional challenges to a DIMT system.
    \item Lack of samples and a unified benchmark: Due to the high annotation cost and different annotation protocals, existing datasets often suffer insufficient samples, inconsistent labels and evaluation metrics, resulting in model performance that are not directly comparable.
    \end{itemize}

\begin{figure}[!t]
\begin{center}
\includegraphics[width=\linewidth]{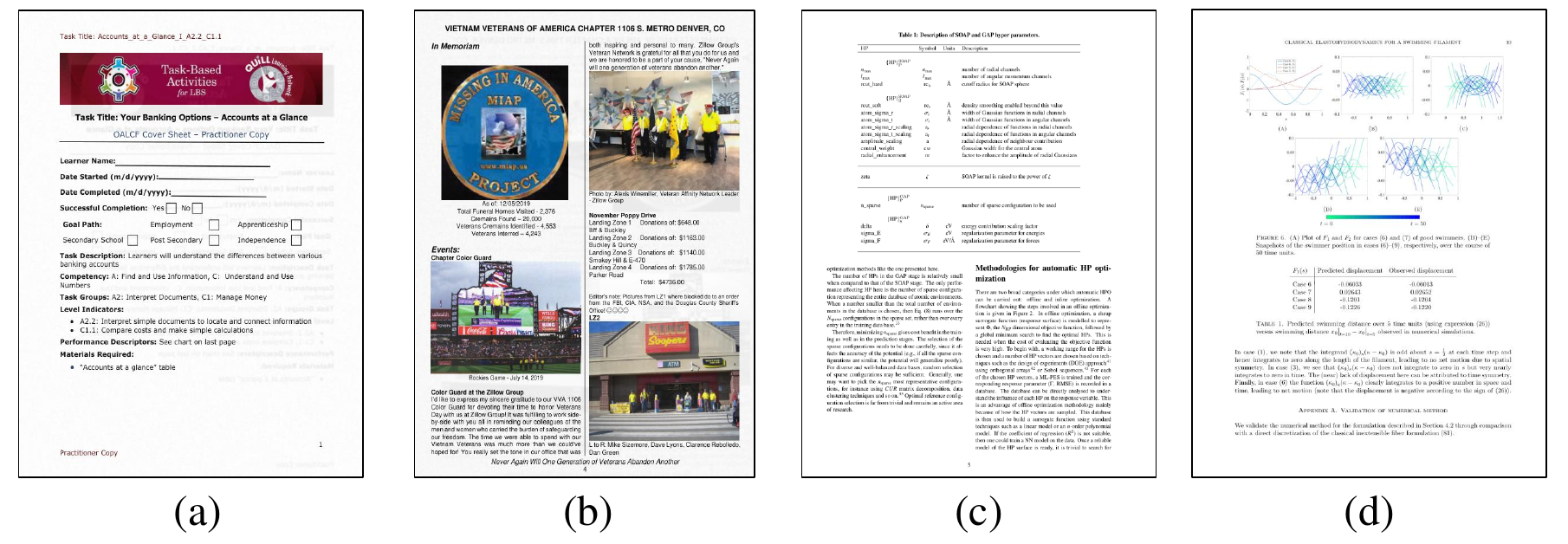}
\end{center}
\caption{Document image examples of the DIMT 2025 challenge. (a)(b): Web document examples in Track 1; (c)(d): arXiv document examples in Track 2.}
\label{fig:example}
\end{figure}

To address these gaps, we introduce the end-to-end Document Image Machine Translation Challenge at ICDAR 2025 (DIMT 2025), the first comprehensive benchmark for document image translation. This challenge offers standardized, large-scale datasets with detailed annotations and a unified evaluation protocol, featuring document images with complex layouts (see Fig.~\ref{fig:example}). Participants are invited to develop models across two complementary tracks:
\begin{itemize}
  \item \textbf{OCR-based:} 
  This track focuses on translating text from OCR-extracted segments. The core task involves accurately reordering these potentially chaotic word-level outputs and translating them into a coherent, structurally sound target-language text that faithfully preserves the original document's semantic content and layout integrity.
  
    \item \textbf{OCR-free:} This track focuses on end-to-end DIMT without any OCR assistance, where participants translate source document images directly into markdown-formatted target translations while handling complex layouts and contextual information.
\end{itemize}

Each track consists of 2 subtasks: small-model (with $\leq$ 1B parameters) and large-model (with $>$ 1B parameters), encouraging innovation in both resource-constrained and large-scale settings. Submissions are evaluated using an end-to-end document-level translation metric, referred to as document-level BLEU \cite{papineni-etal-2002-bleu}. 
By deeply fusing OCR and NLP into a unified multimodal framework and rigorously evaluating on complex layouts, the DIMT 2025 Challenge aims to eliminate brittle preprocessing pipelines and catalyze breakthroughs in multilingual document understanding and translation.

\section{Related Works}
\quad LLMs have significantly impacted Document AI, with developments classifiable by their reliance on OCR: OCR-based methods using text from OCR engines, and OCR-free methods directly processing document images.

\textbf{OCR-based Document AI.} 
OCR-based methods first extract text using an OCR engine, then feed this, often with layout data, to an LLM. The LayoutLM series \cite{layoutlm,xu2021layoutlmv2} was foundational, pre-training models by integrating text, 2D position, and visual embeddings from OCR outputs. LiLT \cite{wang2022lilt} extended this with a language-independent layout transformer.

\textbf{OCR-free Document AI.}
OCR-free methods bypass OCR error propagation by directly interpreting document images. Donut \cite{donut} pioneered this with an end-to-end transformer converting images to structured text. LayoutLMv3 \cite{Huang2022LayoutLMv3PF} advanced multimodal pre-training, capable of OCR-free understanding by treating text detection as a pre-training task using image patch embeddings. More recently, Nougat \cite{nougat} focuses on converting scientific document images into structured Markdown, performing OCR and parsing simultaneously. General-purpose Vision-Language Models (VLMs) like GPT-4o\cite{achiam2023gpt,hurst2024gpt}, InternVL~\cite{dong2024internlm}, LLaVA \cite{zhang2024llava}, and MiniCPM\cite{yao2024minicpm} also show promise in zero-shot OCR-free document understanding by reasoning about text in images. These methods leverage visual and spatial cues but can be computationally intensive and sensitive to visual variations.

While these approaches have significantly advanced the field, existing visually-rich document benchmarks predominantly focus on evaluating coarse-grained understanding. Common tasks include document visual question answering (e.g., DocVQA \cite{docvqa2021}, RDVQA \cite{RDVQA2022}, CS-DVQA \cite{CS-DVQA2022}) or structured/key information extraction (e.g., VisualMRC \cite{Visualmrc2021}, DUDE \cite{DUDE2023}). Even recent, large-scale benchmarks like MMVQA \cite{MMVQA2024}, which provides extensive QA pairs over scientific documents including tables and figures, and DocILE \cite{docile2023}, offering numerous real and synthetic documents for key information and line item recognition in financial/legal domains, primarily assess the ability to locate and interpret specific pieces of information rather than demanding a comprehensive understanding of all textual content.

Additionally, we provide a multilingual dataset with a broader range of categories.  An overview of these datasets is presented in Table~\ref{table:Dataset_description}. Notably, DIT700K \cite{zhang-etal-2025-chaotic} is a large-scale document image translation dataset containing 718,000 images (619,000 in English, 99,000 in Chinese), with multi-level fine-grained labels including word text/box, sentence prefix/order/translation, and document translation, across three translation directions (En→Zh/De, Zh→En). Constructed through an automated pipeline (web crawling, XML/PDF processing, color-coded word-box matching), DIT700K spans diverse disciplines and layout complexities, which are categorized into simple and complex subsets based on the Layout Score.

Furthermore, DoTA \cite{liang-etal-2024-document} introduces a new challenge: Document Image Machine Translation to Markdown (DIMT2Markdown). This challenge aims to translate source document images with long-context and complex layout structures into markdown-formatted target translations. The accompanying dataset consists of 126,345 paired images, English markdown files, and their corresponding translations in Chinese, French, and German. The dataset is split into 124,338 images for training, 1,004 for validation, and 1,003 for testing.


\begin{table*}[htbp]
\centering
\caption{Existing Visually-rich Document Image Datasets.} 
\label{table:Dataset_description}
\setlength{\tabcolsep}{1.2mm} 
\renewcommand{\arraystretch}{1.2}
\scalebox{0.88}{
\begin{tabular}{l c c c l c} 
\hline
\textbf{Dataset} & \textbf{Images} & \textbf{Language} & \textbf{Granularity} & \textbf{Document Type} & \textbf{Year} \\
\hline
DocVQA \cite{docvqa2021} & 12,767 & English & Word & Industrial Reports & 2021 \\
VisualMRC \cite{Visualmrc2021} & 10,197 & English & Word & Website & 2021 \\
RDVQA \cite{RDVQA2022} & 8,514 & English & Word & Data Analysis Report & 2022 \\
CS-DVQA \cite{CS-DVQA2022} & 600 & English & Word & Industry Documents & 2022 \\
DUDE \cite{DUDE2023} & 28,709 & English & Word & Cross-Domain & 2023 \\
DITrans \cite{zhang2023layoutdit} & 1,675 & English & Word & Report, News, Ad. & 2023 \\ 
MMVQA \cite{MMVQA2024} & 30,239 & English & Word & Academic Paper & 2024 \\
DIT700K \cite{zhang-etal-2025-chaotic} & 718,000 & En, Zh & Word & Web Documents & 2025 \\
DoTA \cite{liang-etal-2024-document} & 126,345 & 4 languages & Page & Academic Paper & 2024 \\
\hline
\end{tabular}
} 
\end{table*}

\section{Benchmark Description}
\subsection{Organization}

\quad ICDAR 2025 competition on DIMT is organized by a team from the Institute of Automation, Chinese Academy of Sciences (CASIA).
The DIMT2025 competition is presented on the official website~\footnote{\url{https://cip-documentai.github.io/}}, which provides datasets download links, baseline code download links, and submission pages for uploading results via CodaLab. We received substantial support from the CodaLab web team.

Based on the input type, we release two tracks: 1) OCR-based DIMT, where model inputs include the image and its OCR results (word and word bounding box); 2) OCR-free DIMT, where model input is the image. Both tracks are given English document images and are required to translate them to Chinese.

\subsection{Track 1. OCR-based DIMT} 

\begin{figure}
    \centering
    \includegraphics[width=\linewidth]{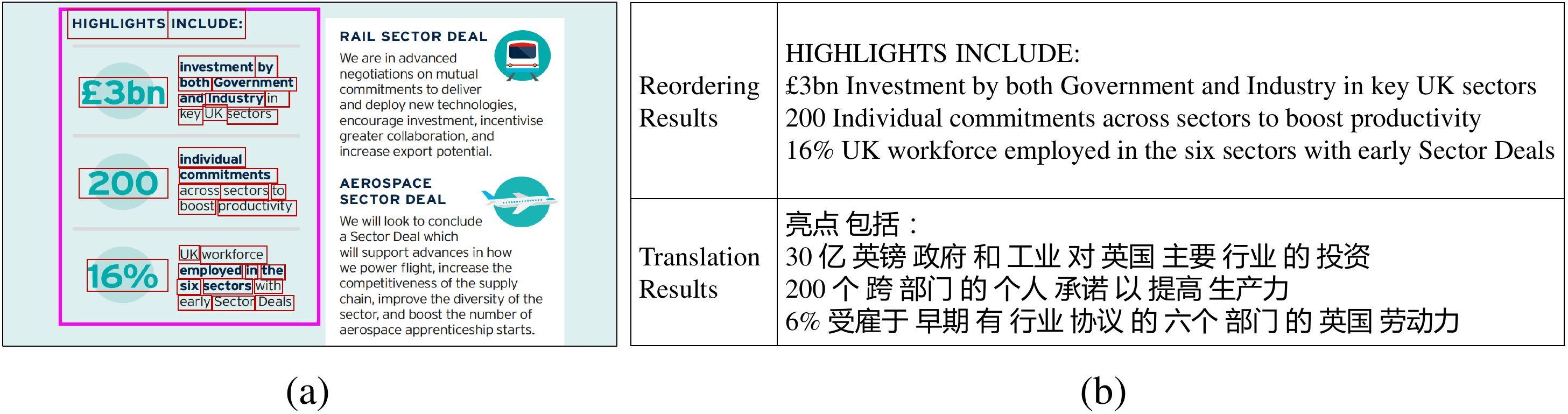}
    \caption{An example of input and output for Track 1 is shown in (a). The \textcolor{purple}{purple} rectangular box highlights the text areas that need to be reordered and translated, with the words and their corresponding bounding boxes (indicated by the \textcolor{red}{red} boxes) serving as the model's input. For clarity, only the area within the rectangular box is illustrated, but in practice, the entire page is considered. (b) The results of reordering and translating the text in the rectangular box area.}
    \label{fig:dit700k_dataset}
\end{figure}

\quad This track focuses on translating text from document images where OCR results (containing words and word bounding boxes) are provided, as shown in Fig.\ref{fig:dit700k_dataset} . The task aims to generate an intact, ordered target translation from the raw, often chaotic OCR-extracted words in an end-to-end manner. Participants are required to handle misordered, fragmented, or disjointed OCR outputs and accurately reorder them while translating the text into the target language. The goal is to produce a coherent and structured target-language text that reflects the original document's content and layout, ensuring the translation maintains both meaning and order.

\textbf{Track 1.1 OCR-based DIMT-LLM}: aims to evaluate the performance of LLM-based methods on the DIMT task. In this sub-track, participants must use large language models (LLMs) with over 1 billion parameters to achieve the OCR-based DIMT task. Open-source LLMs can be utilized, and participants are allowed to fine-tune these models to improve performance. The number of parameters in the model must be specified in the submitted reports used during inference.

\textbf{Track 1.2 OCR-based DIMT-Small}: aims to evaluate the performance of small model-based methods on the DIMT task. In this sub-track, participants are only allowed to use small models, where the number of parameters is fewer than 1 billion, to achieve the OCR-based DIMT task. Participants must focus on optimizing these smaller models for accurate translation and reordering. The number of parameters in the model must be specified in the submitted reports used during inference.

\subsection{Track 2. OCR-free DIMT} 
\begin{figure}
    \centering
    \includegraphics[width=0.8\linewidth]{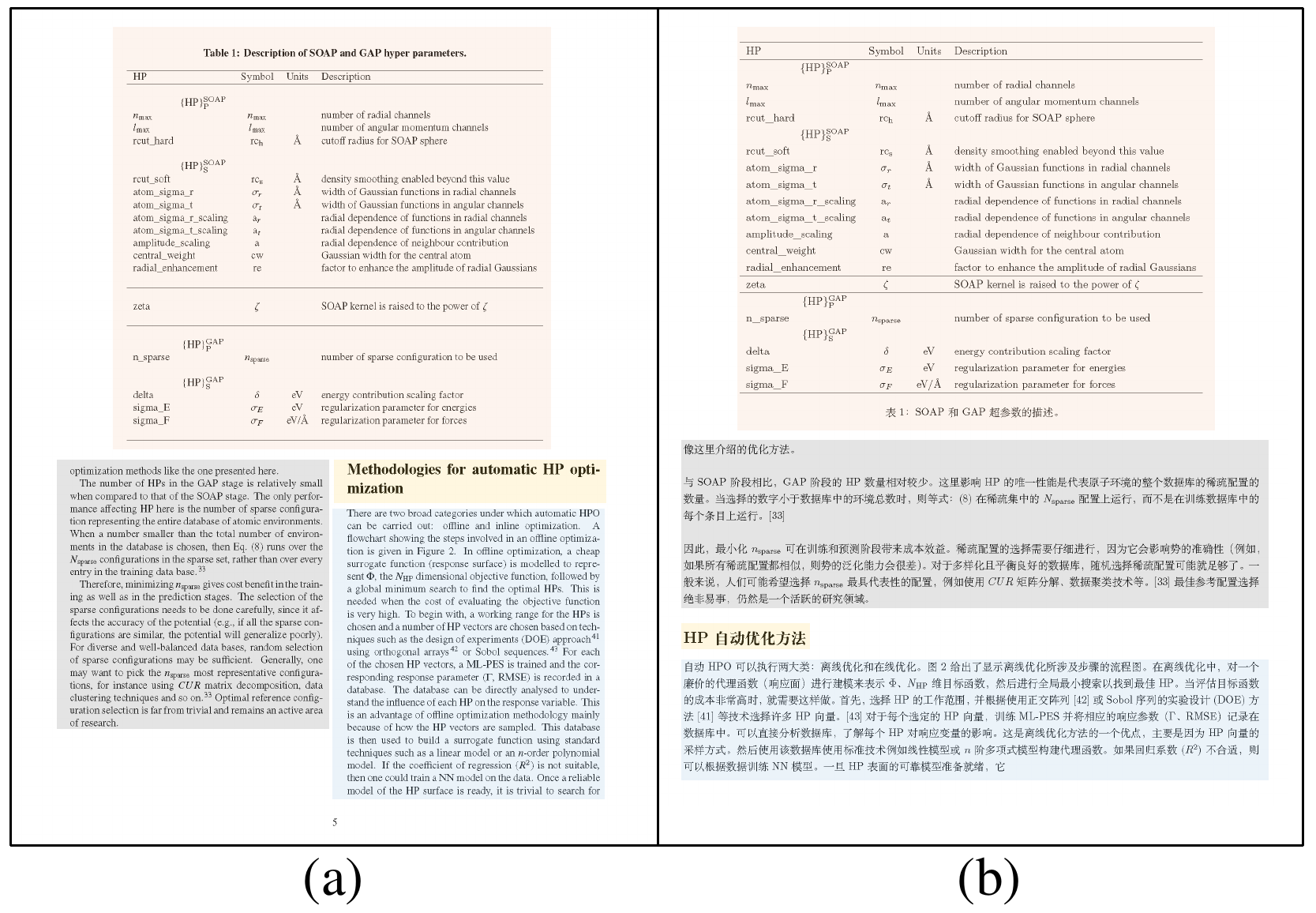}
    \caption{An input and output example of Track 2. (a) is the original document image. (b) is the translated text in markdown format after rendering. The blocks with the same color represent the corresponding original text and translated text.}
    \label{fig:dota_dataset}
\end{figure}

\quad This track focuses on an OCR-free approach to DIMT, where participants are tasked with translating a source document image directly into a markdown-formatted target translation without any OCR assistance in an end-to-end way, as shown in Fig.~\ref{fig:dota_dataset}. In the OCR-free scene, participants must handle the complex layout and contextual information directly from the image, delivering a coherent and accurate translation in the target language. The final submission must follow markdown format reflecting the content and layout of the original document.
\vspace{-10pt}

\textbf{Track 2.1 OCR-free DIMT-LLM}: aims to evaluate the performance of LLM-based methods on the OCR-free DIMT task. In this sub-track, participants must use large language models (LLMs) with over 1 billion parameters to achieve the OCR-free DIMT task. Open-source LLMs can be utilized and fine-tuned to handle complex layouts and long context translation. Participants must specify the number of parameters in the model used during inference in their submitted reports.

\textbf{Track 2.2 OCR-free DIMT-Small}: aims to evaluate the performance of small model-based methods on the OCR-free DIMT task.
In this sub-track, participants are only permitted to use small models with fewer than 1 billion parameters to achieve the OCR-free DIMT task. Participants must work within these constraints, focusing on optimizing smaller models to handle complex layouts and long context. The number of parameters in the model must be specified in the submitted reports used during inference.

\section{Dataset}
\quad In this competition, we have assembled a dataset of over 42,400 document pages drawn from two domains: web documents and scientific articles, with a held-out test set of 1,000 pages. Detailed statistics of the competition dataset are summarized in Table~\ref{tab:dataset-statistics}. The following provides a brief description of the datasets used in this competition. For more comprehensive details, please refer to our previous work \cite{liang-etal-2024-document,zhang-etal-2025-chaotic}.

\begin{itemize}
\item \textbf{DIMT-WebDoc-300K} dataset for Track 1: This dataset includes a training set of 300,000 images and a validation set of 1,000 images, all derived from publicly available web documents. Each image is accompanied by the corresponding OCR results, which include word-level text and bounding boxes, as well as the word-level reading order index, sentence-level translations, and document-level translations.
\item \textbf{DIMT-arXiv-124K} dataset for Track 2: This dataset includes a training set of 124,000 images and a validation set of 1,000 images. These images are extracted from PDF and LaTeX files crawled from Arxiv. Each image is paired with corresponding source language text and target language text in markdown format.
\end{itemize}

\begin{table}[h]
\centering
\caption{Statistical data information of the DIMT 2025 challenge.}
\begin{tabular}{l|l|ccc}
\hline
\multicolumn{1}{c|}{\multirow{2}{*}{Track}} & \multicolumn{1}{c|}{\multirow{2}{*}{Dataset}} & \multicolumn{3}{c}{\# of Examples} \\
\multicolumn{1}{c|}{}                  & \multicolumn{1}{c|}{}                              & Train     & Valid     & Test     \\ \hline
Track 1                                & DIMT-WebDoc-300K                                            & 300K      & 1K        & 1K       \\
Track 2                                & DIMT-arXiv-124K                                               & 124K      & 1K        & 1K       \\ \hline
\end{tabular}
\label{tab:dataset-statistics}
\end{table}

\section{Evaluation Protocol}
\quad We adopt the evaluation metric used in DoTA \cite{liang-etal-2024-document} and DIT700K \cite{zhang-etal-2025-chaotic}, employing document-level BLEU \cite{papineni-etal-2002-bleu} for both tracks.
Participants are asked to treat the final translation output of an entire image as a single text string. Chinese segmentation is performed using Jieba~\footnote{\url{https://github.com/fxsjy/jieba}}, followed by BLEU-4 score computation between the predicted translation and the reference text using the NLTK toolkit~\footnote{\url{https://github.com/nltk/nltk}}.

\section{Submissions and Results}
\quad The competition attracted 69 participants and 27 valid submissions from both academia and industry. Participant statistics are provided in Table~\ref{tab:participation_statistic}. During the competition, we established an evaluation platform on CodaLab, which automatically computed BLEU scores for participants' submissions in real-time and updated the leaderboard accordingly. As each new submission would overwrite the previous one, we reminded participants to upload their best final results before the deadline. The winners for each track were determined based on the highest BLEU scores. The complete leaderboard for all tracks is available on the official website~\footnote{\url{https://cip-documentai.github.io/}}.

\begin{table*}[htbp]
\label{tab:participation_statistic}
\caption{The statistics of the competition submission }
\centering
\begin{tabular}{l|c|c|c}
\hline
Track                                   & Subtrack & Participate & Submission \\ \hline
\multirow{3}{*}{Track 1 OCR-based DIMT} & LLM      & 18          & 6          \\
                                        & Small    & 16          & 7          \\
                                        & Total    & 34          & 13         \\ \hline
\multirow{3}{*}{Track 2 OCR-free DIMT}  & LLM      & 20          & 9          \\
                                        & Small    & 15          & 5          \\
                                        & Total    & 35          & 14         \\ \hline
\end{tabular}

\end{table*}

After the competition, we required participants to specify the number of model parameters and whether any additional training data was used in their technical reports. This was in accordance with the competition rules, which permitted the use of pre-trained models but restricted training to the provided dataset. The number of model parameters determined each team’s subtrack classification. Ultimately, all submitted technical reports met these requirements.

\subsection{Track 1.1 OCR-based DIMT-LLM}
\quad In this section, we present the results and methods employed by the submissions for Track 1.1 OCR-based DIMT-LLM. Models with more than 1 billion parameters are used to achieve OCR-based end-to-end document image machine translation. In this setup, the input consists of the raw OCR results from a document, and the output is the document-level translation, including the reading order. The reported results are selected from the top five entries in the leaderboard, based on the intersection with the submitted method reports, as shown in Table~\ref{results1.1}.

\begin{table*}[htbp]
\caption{The results of Track 1.1 OCR-based DIMT-LLM }
\label{results1.1}
\centering
\small
\resizebox{\linewidth}{!}{
\begin{tabular}{l l l l l  }
\hline
Rank & Team Name  & Team Members   & Institute   & BLEU   \\ \hline
Baseline  & - & - & - & 26.34 \\ \hline
1    & Hw-tsc     & \begin{tabular}[c]{@{}l@{}}Zhanglin Wu, Tengfei Song, Ning Xie,\\ Weidong Zhang, Pengfei Li, Shuang Wu,\\ Chong Li, Junhao Zhu, and Hao Yang\end{tabular} & Huawei Translation Service Center                                                                                                               & 70.48 \\ \hline
2    & Lucky Star & \begin{tabular}[c]{@{}l@{}}Zhentao Guo, Zining Wang, Tongkun Guan,\\ Hao Sun, Chen Duan, and Kai Zhou\end{tabular}                                        & \begin{tabular}[c]{@{}l@{}}Beijing Institute of Technology,\\ Meituan,Shanghai Jiao Tong University,\\ Chinese Academy of Sciences\end{tabular} & 34.35 \\ \hline
3    & Abantika Bose    & Abantika Bose                                                                                                                                             & \begin{tabular}[c]{@{}l@{}}Friedrich-Alexander-Universität\\ Erlangen-Nürnberg\end{tabular}                                                  & 10.64  \\ \hline
\end{tabular}}
\end{table*}

\textbf{Baseline method.} We provide a baseline in which Qwen2-VL-7B \cite{qwen2-vl} is employed to directly generate the complete translated text based on the input document image and its corresponding OCR results.

\textbf{1st ranking method.} "Hw-tsc" team primarily utilizes the InternVL2.5-8B-MPO framework \cite{internvl2_5}, which integrates multi-task learning with a perceptual chain-of-thought training approach. This enables the model to process both visual layouts and linguistic content concurrently. During inference, the system further applies minimum Bayesian decoding and post-processing techniques to optimize the quality of the generated output.

\textbf{2nd ranking method.} "Lucky Star" team's model architecture employs an Encoder-Decoder framework, comprising a LayoutLM-based \cite{layoutlm} Encoder and a Qwen2.5-7B-based \cite{qwen2_5} Decoder. The LayoutLM Encoder processes both textual content and the coordinates of text boxes from document images. By leveraging the Transformer mechanism, it integrates textual content with visual layout features to produce rich contextual and layout-aware representations. These representations are subsequently passed through a latent space to the Qwen2.5-7b Decoder. The Qwen2.5-7b Decoder receives the context-rich representations generated by the Encoder and, based on its advanced Transformer-based language modeling capabilities, is fine-tuned to generate target-language text sequences. This process ensures that the translated output is coherent, fluent, and natural.

\textbf{3rd ranking method.} "Abantika Bose" team splits the DIMT task into two sequential stages: Stage 1: Reordering: Llama-4-Maverick-17B-128E (402 B) \cite{llama} via few-shot in-context prompting, no fine-tuning. Stage 2: Translation. Reordered English text from either expert is translated into Chinese using Helsinki-NLP/opus-mt-en-zh (77.9 M) \cite{opus-mt}, followed by Jieba segmentation.

\subsection{Track 1.2 OCR-based DIMT-Small}
\quad In this section, we present the results and methods employed by the submissions for Track 1.2 OCR-based DIMT-Small. In this track, the input and output are consistent with Track 1.1, with the only difference being that the models used in this track must have fewer than 1 billion parameters. The aim of this track is to explore the potential of smaller models in comparison to large models. The reported results are selected from the top five entries in the leaderboard, based on the intersection with the submitted method reports, as shown in Table~\ref{results1.2}.

\begin{table*}[htbp]
\caption{The results of Track 1.2 OCR-based DIMT-Small }
\label{results1.2}
\centering
\small
\resizebox{\linewidth}{!}{
\begin{tabular}{l l l l l}
\hline
Rank & Team Name  & Team Members                                                                                                                                              & Institute                                                                                                                                    & BLEU  \\ \hline
Baseline & - & - & - & 38.81 \\ \hline
1    & Hw-tsc     & \begin{tabular}[c]{@{}l@{}}Zhanglin Wu, Tengfei Song, Ning Xie,\\ Weidong Zhang, Pengfei Li, Shuang Wu,\\ Chong Li, Junhao Zhu, and Hao Yang\end{tabular} & Huawei Translation Service Center                                                                                                            & 66.16 \\ \hline
2    & CMC ATI    & \begin{tabular}[c]{@{}l@{}}Doan Kien Thai, An Nguyen Thai,\\ and Vu Son Lam Nguyen\end{tabular}                                                           & CMC ATI                                                                                                                                      & 39.73 \\ \hline
3    & sunmk      & Mai Sun, Jiazi Wang, and Long Yuan                                                                                                                        & Beijing Jiaotong University                                                                                                                  & 37.06 \\ \hline
4    & Lucky Star & \begin{tabular}[c]{@{}l@{}}Zhentao Guo, Zining Wang, Tongkun Guan,\\ Hao Sun, Chen Duan, and Kai Zhou\end{tabular}                                        & \begin{tabular}[c]{@{}l@{}}Beijing Institute of Technology,\\Meituan,\\Shanghai Jiao Tong University,\\Chinese Academy of Sciences\end{tabular} & 33.61 \\ \hline
5    & Abantika Bose     & Abantika Bose                                                                                                                                             & \begin{tabular}[c]{@{}l@{}}Friedrich-Alexander-Universität\\ Erlangen-Nürnberg\end{tabular}                                                  & 9.40  \\ \hline
\end{tabular}}
\end{table*}

\textbf{Baseline method.} We construct an encoder-decoder model based on LayoutLM \cite{layoutlm} and a Transformer decoder, and train it on the provided training data for 10,000 steps.

\textbf{1st ranking method.} The Hw-tsc team utilized the same framework as in the Track 2.1 OCR-based 
 DIMT-LLM track, but with a smaller pretrained version, InternVL2.5-1B \cite{internvl2_5}. This approach enabled the team to secure first place in the competition.

\textbf{2nd ranking method.} "CMC ATI" team's pipeline includes 2 models for distinct tasks: Text Ordering and Machine Translation. The outputs from Text Ordering Model will be served as the input to the Machine Translation model. In detail, they used the LayoutLMv3 \cite{layoutlmv3} model for the reordering model due to its pre-trained English ability and the Qwen2.5-0.5B-Instruct \cite{qwen2_5} translation model due to the fact that the predominant languages trained on this model are reported to be English and Chinese. In both models, they performed supervised fine-tuning on the competition dataset to enhance the
specialized ability in each task.

\textbf{3rd ranking method.} "sunmk" team proposes a layout-aware encoder-decoder architecture for document-level sequence generation tasks, where the input consists of recognized words along with their corresponding bounding boxes. The encoder leverages LayoutLM \cite{layoutlm} to incorporate spatial layout information, while the decoder is based on BERT-base \cite{bert} and generates textual sequences in an auto-regressive manner. The encoder and decoder are connected through a Transformer-based cross-attention fusion module.

\subsection{Track 2.1 OCR-free DIMT-LLM}
\quad In this section, we present the results and methods employed by the submissions for Track 2.1 OCR-free DIMT-LLM. In this setup, the model is fully end-to-end, with the input being a document image and the output in markdown format, including structural information. Models with more than 1 billion parameters are used to achieve the task of Track 2.1. The reported results are selected from the top five entries in the leaderboard, based on the intersection with the submitted method reports, as shown in Table~\ref{results2.1}.

\begin{table*}[htbp]
\caption{The results of Track 2.1 OCR-free DIMT-LLM }
\centering
\small
\resizebox{\linewidth}{!}{
\begin{tabular}{l l l l l}
\hline
Rank & Team Name     & Team Members                                                                                                                                              & Institute                                                                                                                                      & BLEU  \\ \hline
Baseline  & - & - & - & 48.63 \\ \hline
1    & Hw-tsc        & \begin{tabular}[c]{@{}l@{}}Zhanglin Wu, Tengfei Song, Ning Xie,\\ Weidong Zhang, Pengfei Li, Shuang Wu,\\ Chong Li, Junhao Zhu, and Hao Yang\end{tabular} & Huawei Translation Service Center                                                                                                              & 60.78 \\ \hline
2    & 360ailab      & Junhui Yu                                                                                                                                                 & 360 AI Research Institute                                                                                                                      & 57.78 \\ \hline
3    & Lucky Star    & \begin{tabular}[c]{@{}l@{}}Zhentao Guo, Zining Wang, Tongkun Guan,\\ Hao Sun, Chen Duan, and Kai Zhou\end{tabular}                                        & \begin{tabular}[c]{@{}l@{}}Beijing Institute of Technology,\\Meituan,\\Shanghai Jiao Tong University,\\Chinese Academy of Sciences\end{tabular} & 44.41 \\ \hline
\end{tabular}}
\end{table*}

\textbf{Baseline method.} We provide participants with a baseline model, where LoRA fine-tuning \cite{lora} is applied to adapt Qwen2-VL-7B\cite{qwen2-vl} on the provided training data for 1,000 steps.

\textbf{1st ranking method.}  "Hw-tsc"  team's OCR-free-LLM approach integrates OCR-based and OCR-free translation tasks within a single framework, eliminating the need for separate pipelines. This approach is built upon the InternVL2.5-8B-MPO framework \cite{internvl2_5}.

\textbf{2nd ranking method.} "360ailab" team proposes a robust end-to-end document translation method based on the Qwen2.5-VL multimodal large language model \cite{qwen2_5-vl}. Additionally, through adversarial training, they apply adversarial training algorithms related to the visual encoder of the large model, such as Projected Gradient Descent and Fast Gradient Method, to enhance the model’s robustness and improve its generalization ability.

\textbf{3rd ranking method.} "Lucky Star" team's methods are mainly based on CLIP \cite{clip} and Qwen2.5-7B \cite{qwen2_5}. The CLIP encoder is responsible for processing the input document images, extracting both visual and textual content features. This encoding phase converts complex page layouts into a structured set of representations, bridging the gap between visual appearance and semantic content. Subsequently, these representations are fed into the Qwen2.5-7B decoder. This decoder utilizes its expansive language model to transform the encoded data into target language translations, ensuring context-aware and semantically accurate outputs. The integration between these components is facilitated through a tailored interface that harmonizes visual and textual data flow, optimizing the translation process.

\subsection{Track 2.2 OCR-free DIMT-Small}
\quad In this section, we present the methods employed by the submissions for Track 2.2 OCR-free DIMT-Small, as shown in Table~\ref{results2.1}, highlighting the diverse approaches used to address the challenges of OCR-free document image machine translation. The reported results are selected from the top six entries in the leaderboard, based on the intersection with the submitted method reports.

\begin{table*}[htbp]
\caption{The results of Track 2.2 OCR-free DIMT-Small }
\label{results2.1}
\centering
\small
\resizebox{\linewidth}{!}{
\begin{tabular}{l l l l l}
\hline
Rank & Team Name   & Team Members                                                                                                                                                & Institute                                                                                                                                          & BLEU  \\ \hline
Baseline  & - & - & - & 27.03 \\ \hline
1    & Intime \& HY & \begin{tabular}[c]{@{}l@{}}Gengluo Li, Huawen Shen, Chengquan Zhang,\\ Xingyu Wan, Pengyuan Lyu, Liang Wu,\\ Zhuohao Chen, Han Hu, and Yu Zhou\end{tabular} & \begin{tabular}[c]{@{}l@{}}Chinese Academy of Sciences,\\ Tencent,\\ Nankai University\end{tabular}                                                & 59.96 \\ \hline
2    & Hw-tsc      & \begin{tabular}[c]{@{}l@{}}Zhanglin Wu, Tengfei Song, Ning Xie,\\ Weidong Zhang, Pengfei Li, Shuang Wu,\\ Chong Li, Junhao Zhu, and Hao Yang\end{tabular}   & Huawei Translation Service Center                                                                                                                  & 59.56 \\ \hline
3    & Blue Jays   & \begin{tabular}[c]{@{}l@{}}Cameron Carpenter, David Etter, Matt Lee,\\ Paul McNamee, Kevin Duh, and Kenton Murray\end{tabular}                              & Johns Hopkins University                                                                                                                           & 58.41 \\ \hline
4    & Lucky Star  & \begin{tabular}[c]{@{}l@{}}Zhentao Guo, Zining Wang, Tongkun Guan,\\ Hao Sun, Chen Duan, and Kai Zhou\end{tabular}                                          & \begin{tabular}[c]{@{}l@{}}Beijing Institute of Technology,\\ Meituan,\\ Shanghai Jiao Tong University,\\ Chinese Academy of Sciences\end{tabular} & 44.41 \\ \hline
\end{tabular}}
\end{table*}

\textbf{Baseline method.} We provide the OCR-free DIMT small baseline model, which uses Nougat's encoder \cite{nougat} along with a pre-trained translation decoder. A Transformer-based model is pre-trained on the DoTA dataset for the text machine translation task. Subsequently, an encoder-decoder model is constructed using Nougat's encoder \cite{nougat} and the pre-trained translation decoder. The model is further fine-tuned for 20,000 steps on the provided dataset.

\textbf{1st ranking method.} "Intime \& HY" team combines document parsing-based self-filtering with reinforcement learning to enhance the performance of small model-based OCR-free document image machine translation. The team employs a self-developed model named HYOCR-1B as the backbone and further trains it on English-Chinese document data to enhance structured document parsing capabilities. They also use the DPO reinforcement learning strategy to optimize the model's outputs, reducing hallucinations and improving translation accuracy.

\textbf{2nd ranking method.} "Hw-tsc" team's methods are mainly based on InternVL2.5-1B-MPO \cite{internvl2_5}. This framework integrates multi-task learning with perceptual chain-of-thought training approach, enabling the model to simultaneously comprehend visual layouts and linguistic content. During the inference phase, the system further employs minimum Bayesian decoding and post-processing strategies to optimize output quality.

\textbf{3rd ranking method.} "Blue Jays" team trained a small end-to-end vision-language model, which follows a three-part design consisting of a Vision Transformer image encoder \cite{vit}, an MLP projector, and an auto-regressive LLM decoder \cite{transformer}.

\section{Discussion}
\quad In this section, we analyze participation trends, performance gaps, and the methods used in the submission reports, as shown in Table~\ref{tab:method-compare}.

\subsection{Participation Trends and Performance Gap}

\quad \textbf{Participation Trends}. The competition provided a unique opportunity to explore the interest in both OCR-based and OCR-free paradigms. Participation in the OCR-based and OCR-free tracks was nearly identical, with 34 teams in the OCR-based track and 35 in the OCR-free track, reflecting similar levels of interest in both approaches. However, a notable trend emerged: the LLM tracks (both OCR-based and OCR-free) consistently had more participants than their small-model counterparts. Among them, the OCR-free-LLM track attracted the highest number of submissions, highlighting the growing confidence in using large-scale models for OCR-free document image translation, a historically more challenging task.

\textbf{Performance Gap}. The performance gap between OCR-based and OCR-free models remains substantial. OCR-based models consistently delivered better results, underscoring the reliability of OCR methods in extracting text from document images. In contrast, OCR-free models are still catching up and face greater challenges when translating documents without OCR support. However, the progress made by OCR-free models, especially in large-scale tracks, suggests that these models will eventually close the performance gap with OCR-based systems as they continue to improve.

\begin{table}[]
\caption{The method reports submitted by the participating teams}
\resizebox{\textwidth}{!}{
\begin{tabular}{cccccccc}
\hline
\multicolumn{8}{c}{OCR-based}                                                                                                                                                                                                                                                                   \\ \hline
     Model Type           & Team Name                            & Method                                     &               & SFT                       & DPO                       & Parameters                 & \begin{tabular}[c]{@{}c@{}}Extra \\ Data\end{tabular} \\ \hline
                               & \cellcolor[HTML]{C0C0C0}hw-tsc       & \cellcolor[HTML]{C0C0C0}InternVL2.5-8B-MPO & \cellcolor[HTML]{C0C0C0}70.48 & \cellcolor[HTML]{C0C0C0}$\checkmark$ & \cellcolor[HTML]{C0C0C0}$\checkmark$ & \cellcolor[HTML]{C0C0C0}8B & \cellcolor[HTML]{C0C0C0}                              \\
                               & Lucky Star                           & LayoutLM+qwen2.5-7B                        & 34.35         & $\checkmark$                        & ×                         & 7.3B                       &                                                       \\
\multirow{-3}{*}{LLMs}         & Abantika Bose                              & LLaMA-402B                                 & 10.64                         & ×                         & ×                         & 402B                       &                                                       \\ \hline
                               & \cellcolor[HTML]{C0C0C0}hw-tsc       & \cellcolor[HTML]{C0C0C0}InternVL2.5-1B-MPO & \cellcolor[HTML]{C0C0C0}66.16 & \cellcolor[HTML]{C0C0C0}$\checkmark$ & \cellcolor[HTML]{C0C0C0}$\checkmark$ & \cellcolor[HTML]{C0C0C0}1B & \cellcolor[HTML]{C0C0C0}                              \\
                               & tadkt                                & LayoutLMv3+Qwen2.5-0.5B-Instruct           & 39.73          & $\checkmark$                      & ×                         & 633M                       &                                                       \\
                               & sunmk                                & LayoutLM+BERT                              & 37.06                         &$\checkmark$                     & ×                         & 211M                       &                                                       \\
                               & Lucky Star                           & LayoutLM+RobeERTa-largge                   & 33.61                         &$\checkmark$                      & ×                         & 674M                       & $\checkmark$                                                   \\
\multirow{-5}{*}{Small Models} & Abantika Bose                              & LayoutT5-Reorder                           & 9.4                           & ×                         & ×                         & 264M                       &                                                       \\ \hline
\multicolumn{8}{c}{OCR-free}                                                                                                                                                                                                                                                                    \\ \hline
                               & \cellcolor[HTML]{C0C0C0}hw-tsc       & \cellcolor[HTML]{C0C0C0}InternVL2.5-8B-MPO & \cellcolor[HTML]{C0C0C0}60.78 & \cellcolor[HTML]{C0C0C0}$\checkmark$ & \cellcolor[HTML]{C0C0C0} $\checkmark$     & \cellcolor[HTML]{C0C0C0}8B & \cellcolor[HTML]{C0C0C0}                              \\
                               & yujunhuinlp                          & Qwen2.5-VL-7B-Instruct+PGD                 & 57.78                         & $\checkmark$                        & ×                         & 7B                         &                                                       \\
\multirow{-3}{*}{LLMs}         & Lucky Star                           & CLIP+Qwen2.5-7b                            & 44.41                         & $\checkmark$                      & ×                         & 8B                         &                                                       \\ \hline
                               & \cellcolor[HTML]{C0C0C0}Intime \& HY & \cellcolor[HTML]{C0C0C0}HYOCR-1B       & \cellcolor[HTML]{C0C0C0}59.96 & \cellcolor[HTML]{C0C0C0}$\checkmark$ & \cellcolor[HTML]{C0C0C0}$\checkmark$ & \cellcolor[HTML]{C0C0C0}1B & \cellcolor[HTML]{C0C0C0}                              \\
 & hw-tsc              & InternVL2.5-1B-MPO           & 59.56           & $\checkmark$              & $\checkmark$             & 1B       & $\checkmark$       \\
 & ccarpe18      & SigLip-2+Qwen2.5-0.5B-Instruct             & 58.41     & $\checkmark$           & ×        & 927M      &$\checkmark$      \\
\multirow{-4}{*}{Small Models} & Lucky Star     & Donut+Nougat            & 44.41                         & $\checkmark$                       & ×                         & 500M                       &                                                       \\ \hline
\end{tabular}
}
\label{tab:method-compare}
\end{table}
\subsection{Analysis of Participation Methods}

\quad\textbf{OCR-free vs OCR-based Approaches.} OCR-based models consistently outperformed OCR-free models. OCR-based methods leverage well-established OCR technologies to extract text from document images, providing a solid foundation for subsequent translation. These models benefit from decades of OCR algorithm development, which enables robust text extraction even in documents with complex layouts (e.g., multiple columns, tables, footnotes). This makes OCR-based solutions particularly reliable when translating documents that require precise text extraction.

Although OCR-free models did not outperform OCR-based models, they showed considerable progress, particularly in handling document images without OCR. Models like InternVL2 and Qwen2.5 performed well in OCR-free tracks, indicating that OCR-free models are advancing in their ability to handle multimodal document translation tasks. The performance gap between OCR-based and OCR-free models is narrowing, suggesting that as OCR-free models continue to evolve, they will become more competitive with traditional OCR-based systems.

\textbf{Large vs Small Models.} A key takeaway from the competition is the clear advantage of larger models in handling complex document image translation tasks. Models like InternVL2.5-8B-MPO consistently outperformed smaller models in both the OCR-based-LLM and OCR-free-LLM tracks. For example, InternVL2.5-8B-MPO achieved a score of 70.48 in the OCR-based track, significantly higher than the 66.16 achieved by the top small model. Similarly, in the OCR-free track, the top LLM submission reached 60.78, while the best small model reached 59.96. These results underscore the scalability and capacity of larger models to address complex, real-world document image translation tasks involving diverse text types and structural nuances.

Despite the dominance of LLMs in raw performance, smaller models still have value, especially in resource-constrained scenarios or when dealing with less complex document structures. Smaller models like InternVL2-1B and Qwen2.5-0.5B performed competitively, achieving solid results even with limited computational resources. The performance of smaller models suggests they can be highly optimized for specific tasks, achieving good results with a smaller computational footprint.

Interestingly, smaller models, including those with sizes like 1B and 211M, demonstrated that fine-tuning on domain-specific datasets, such as OCR data or translated documents, can significantly boost performance. While LLMs excel at handling the complexity of multimodal tasks, smaller models can still achieve competitive results when appropriately fine-tuned, making them valuable for applications with limited resources or real-time translation needs.

\textbf{Fine-Tuning Strategies.} The competition results underscore the critical role of fine-tuning strategies, particularly Supervised Fine-Tuning (SFT), in enhancing the performance of models for document image translation tasks. SFT was the predominant technique employed across most of the top-performing models, highlighting that fine-tuning on domain-specific data—such as OCR data or document images—is essential for refining model performance. This approach enables models to adapt to the intricacies of document structures and language pairs, leading to substantial improvements in translation accuracy. Notably, Direct Preference Optimization (DPO) \cite{rafailov2023direct} was utilized by both the 1st and 2nd ranked methods in both OCR-free and OCR-based tracks, indicating that DPO holds promising potential in the context of Document Image Machine Translation (DIMT).

\textbf{Pretraining with Specialized Models.} Specialized models, such as LayoutLM, LayoutT5, and SigLip-2, were employed by several teams, demonstrating the continued value of layout-aware techniques in the pretraining process. These models excel at understanding and processing document structures, which is critical for accurate document translation. However, despite their ability to handle complex document layouts, these specialized models did not perform as competitively in the larger tracks as models like InternVL2 or Qwen2.5. This highlights a key challenge in document image translation: while layout-aware models are effective for documents with simpler or more structured layouts, they struggle to match the broader capabilities and performance of general-purpose LLMs that can handle diverse and complex document structures.

In contrast, models with broader pretraining, such as InternVL2 and Qwen2.5, demonstrated superior performance in both OCR-based and OCR-free tracks. These models are better equipped to handle the diversity of document layouts and content types, which are characteristic of complex document image translation tasks.

\section{Conclusion}
\quad We hosted the DIMT competition, which focused on the complex task of translating text from document images, both with and without OCR preprocessing. The competition introduced new datasets and challenges, providing participants with opportunities to explore both OCR-based and OCR-free document image translation. Submissions highlighted significant interest from both academia and industry in the integration of multimodal technologies for document translation, but also revealed the substantial challenges in dealing with complex document structures, such as mixed layouts and non-textual elements, especially in OCR-free scenarios. The results indicate that LLMs show promising potential for overcoming these challenges, but achieving high performance across diverse and real-world document image scenarios remains difficult. Future iterations of this competition could expand on these challenges by introducing even more intricate datasets, encouraging innovation in model architectures and training strategies. This would not only attract further advancements in the intersection of OCR and NLP but also drive progress in the broader field of Document AI, with numerous applications in areas like automatic document processing and cross-lingual information extraction.

\section{Acknowledgements}
This competition is supported by the National Natural Science Foundation of China (No.62476275, No.62336008) and the Young Scientists Fund of The State Key Laboratory of Multimodal Artificial Intelligence Systems (MAIS2024316).
The organizers thank Binyao Xu and Zheng Lian for their significant efforts and support. We are also deeply grateful to the ICDAR 2025 competition chairs (Zhouhui Lian , Michael Coustaty, and Ron Litman), and the CodaLab web team for their crucial assistance and tremendous support throughout the organization of this competition.

\bibliographystyle{unsrt}
\bibliography{mybib}

\end{document}